\documentclass[lettersize,journal]{IEEEtran}
\usepackage{amsmath,amsfonts}
\usepackage{mathrsfs}
\usepackage{bbm}
\usepackage[ruled,vlined]{algorithm2e}
\usepackage{booktabs}
\usepackage{xcolor}
\usepackage{array}
\usepackage[caption=false,font=normalsize,labelfont=sf,textfont=sf]{subfig}
\usepackage{textcomp}
\usepackage{stfloats}
\usepackage{url}
\usepackage{verbatim}
\usepackage{graphicx}
\usepackage{cite}
\usepackage{subcaption}
\usepackage{tabularx}
\usepackage[font=footnotesize,skip=0pt]{caption}
\begin{document}

\title{Evolving Restricted Boltzmann Machine-Kohonen Network for Online Clustering}

\author{J. Senthilnath \IEEEmembership{Senior~Member,~IEEE}, Adithya Bhattiprolu, Ankur Singh, Bangjian Zhou, Min Wu  \IEEEmembership{Senior~Member,~IEEE}, J\'{o}n Atli Benediktsson \IEEEmembership{Fellow,~IEEE}, Xiaoli Li \IEEEmembership{Fellow,~IEEE}
\thanks{This research is supported by the Accelerated Materials Development for Manufacturing Program at the Agency for Science, Technology and Research (A*STAR) via the AME Programmatic Fund by the Agency for Science, Technology and Research under Grant A1898b0043.}
\thanks{J. Senthilnath, Ankur Singh, Bangjian Zhou and Min Wu are with the Department of Machine Intellection, Institute for Infocomm Research, Agency for Science, Technology and Research (A*STAR), 138632 Singapore (e-mails: J\_Senthilnath@i2r.a-star.edu.sg, ankur\_singh@i2r.a-star.edu.sg, BZHOU006@e.ntu.edu.sg, wumin@i2r.a-star.edu.sg).}
\thanks{Adithya Bhattiprolu is with the Department of Electrical \& Electronics Engineering, Birla Institute of Technology and Science, Pilani 333031, India (e-mail: f20170205p@alumni.bits-pilani.ac.in).}
\thanks{J\'{o}n Atli Benediktsson is with the Faculty of Electrical  and Computer Engineering, University of Iceland, 107 Reykjavik, Iceland (e-mail: benedikt@hi.is).}
\thanks{Xiaoli Li is with the Department of Machine Intellection, Institute for Infocomm Research, Agency for Science, Technology and Research (A*STAR), Singapore, School of Computer Science and Engineering, Nanyang Technological University, Singapore and A*STAR Centre for Frontier AI Research, Singapore 138632 (email: xlli@i2r.a-star.edu.sg).} 
\thanks{\noindent This work has been submitted to the IEEE for possible publication. Copyright may be transferred without notice, after which this version may no longer be accessible.}
}

\markboth{Preprint version}%
{Shell \MakeLowercase{\textit{et al.}}: A Sample Article Using IEEEtran.cls for IEEE Journals}


\maketitle

\begin{abstract}
A novel online clustering algorithm is presented where an \textbf{E}volving \textbf{R}estricted \textbf{B}oltzmann \textbf{M}achine (\textbf{ERBM}) is embedded with a \textbf{K}ohonen \textbf{Net}work called \textbf{ERBM-KNet}. The proposed ERBM-KNet efficiently handles streaming data in a single-pass mode using the ERBM, employing a bias-variance strategy for neuron growing and pruning, as well as online clustering based on a cluster update strategy for cluster prediction and cluster center update using KNet. Initially, ERBM evolves its architecture while processing unlabeled image data, effectively disentangling the data distribution in the latent space. Subsequently, the KNet utilizes the feature extracted from ERBM to predict the number of clusters and updates the cluster centers. By overcoming the common challenges associated with clustering algorithms, such as prior initialization of the number of clusters and subpar clustering accuracy, the proposed ERBM-KNet offers significant improvements. Extensive experimental evaluations on four benchmarks and one industry dataset demonstrate the superiority of ERBM-KNet compared to state-of-the-art approaches.
\end{abstract}


\graphicspath{{Figures/}}

\section{Introduction}
\begin{figure}[t]
\begin{center}
\includegraphics[width=1\linewidth]{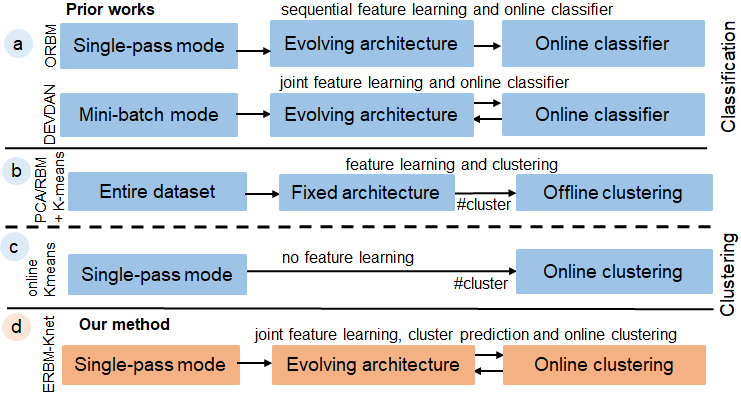} %
\end{center}
\caption{(a) Previous approaches \cite{ashfahani2020devdan,ramasamy2020ORBM} often perform latent representation by storing the entire or mini-batch dataset before applying online classification, (b) Some work \cite{xu2015pca,graphRBM} focuses on batch-mode offline clustering by priori defining the number of clusters, (c) Some work \cite{Vincent2021OKmeans,Hall2011OnlineFCM} focuses on single-pass online clustering in the data space, (d) We unify online latent representation and clustering into a single joint framework, which processes sample-by-sample.}
\label{fig:user_intention}
\vspace{-0.7cm}
\end{figure}

\IEEEPARstart{I}{n} streaming scenarios, the data is obtained sample-by-sample without prior information on their class labels. Therefore, machine learning methods with \textit{batch} mode processing are not applicable. Instead, \textit{online} clustering methods where the model parameters are updated for each encountered data sample, allowing clustering to be performed \textit{on the fly} are needed. The main drawbacks of existing online clustering methods, like the online Kmeans~\cite{barbakh2008online,Vincent2021OKmeans} and online Fuzzy C-Means (FCM)~\cite{hore2008online,Hall2011OnlineFCM} (see Figure~\ref{fig:user_intention}(c)) are: i) clustering performed in the high-dimensional data space, and ii) converge to local optima. To overcome these issues, unsupervised-\textit{feature} learning methods, such as principal component analysis (PCA)-guided Kmeans \cite{xu2015pca} (see Figure~\ref{fig:user_intention}(b)), can be applied. However, PCA operates in batch learning mode that requires the number of transformed dimensions to be predetermined first and is not feasible for nonlinear data distribution. 

Generative models, such as the Restricted Boltzmann Machine (RBM) \cite{hinton2007boltzmann}, Autoencoder \cite{baldi2012autoencoders}, and Denoising Autoencoder (DAE) \cite{vincent2008extracting} offer promising capabilities for capturing nonlinear discrete distributions in a latent space. These models have consistently demonstrated disentangle representation and competitive performance, with minimal performance differences among them \cite{vincent2008extracting, fernandez2023disentangling}. Most of these offline clustering methods (see Figure~\ref{fig:user_intention}(b)) operate in batch/mini-batch mode and fixed architectures, which need to be tuned \cite{gu2021refinements}. Hence, the challenge of these models is to set the number of hidden neurons required to capture the data distribution in the latent space efficiently. It leads to a trial-and-error approach to determine the number of hidden neurons. This study focuses on the evolving RBM (ERBM) architecture, addressing this challenge.

To overcome the issue of architecture tuning the hidden neurons of the generative model, a recent advancement in this field are the energy-based dropout \cite{roder2020energy} and online RBM (ORBM) \cite{ramasamy2020ORBM}. ORBM utilizes a hidden neuron growth strategy by setting a threshold value for online data classification (see Figure~\ref{fig:user_intention}(a)). However, ORBM still has some drawbacks that need to be overcome. Firstly, ORBM only focuses on adding neurons, \textit{without considering the pruning of redundant neurons} that are no longer significant. Secondly, ORBM relies on user-defined thresholds to grow its architecture, which necessitates manual tuning for each dataset. To overcome these limitations, we propose a novel combination of an ERBM for better representation of incoming data and a Kohonen Network (KNet) for online cluster prediction and updates of cluster centers. The ERBM overcomes the threshold dependency issue of ORBM by utilizing a bias-variance strategy. This strategy enables the network to autonomously add and prune neurons based on their significance, resulting in a more efficient latent representation. Furthermore, most existing clustering algorithms need a prior initialization of the number of clusters. In contrast, the proposed ERBM-KNet utilizes the latent representation of ERBM to predict the number of clusters and update the cluster centers in an online mode. Hence, our proposed ERBM-KNet is a fully autonomous online clustering method.

The main contributions of our paper are summarized as three-fold: 
\begin{enumerate}
    \item To the best of our knowledge, this is the first work to propose \textit{novel evolving RBM (ERBM) architecture with growing and pruning neuron for feature-learning} to handle streaming data in a single-pass latent representation. 
    \item Unlike the traditional batch learning-based clustering, \textit{a novel online clustering KNet} is combined with ERBM for cluster prediction and cluster center update.
    \item The experimental results demonstrate that our proposed ERBM-KNet performs better than offline/online state-of-the-art clustering methods across four benchmarks and semiconductor industry wafer defect map data to facilitate accurate online cluster prediction.
\end{enumerate}

The rest of the paper is organized as follows: Section II consists of related works. Section III discusses the proposed ERBM-KNet with theoretical insights. Section IV contains an experimental analysis of 5 image datasets. The paper is concluded in Section V. 

\section{Related works}\label{Literature}
In this section, we briefly introduce the existing autonomous online learning and the generative online learning that are related to our work.

\subsection{Online learning} 
Over the last decade, data streams have become increasingly popular in various real-world applications such as analyzing customer log files from mobile devices and the Web, e-commerce purchases, financial trading, and geospatial image analysis. Hence, they have become an active research topic recently. As a result, several online learning methods have been proposed \cite{golab2003issue,Datastreams1} to extract useful features from incoming data streams. Online learning has become an interesting area of research due to its ability in data streaming and memory-efficient dynamic modeling. In real-world scenarios, it is often necessary to determine the relationship between samples in the data stream. Determining this relationship can help to provide new knowledge and insights and draw new reasoning and inference. Several studies exist on online learning with fixed architecture while only a few work is available on evolving architecture. Also, these studies during training evolve single-layer network architecture for online classification problems like metacognitive learning \cite{babu2016meta,Senthil2022metacog}, and bias-variance strategy \cite{ashfahani2021biasvar}. However, these models are less efficient for complex nonlinear data distributions in high-dimensional feature spaces and need label information. To overcome these issues generative models are popularly used for unsupervised feature learning with a fixed architecture. This study focus on bias-variance strategy to evolve architecture for unsupervised feature learning.

\subsection{Generative online learning} 
Here we address the challenge related to learning complex nonlinear data distribution with high-dimensional feature space in an online fashion and without label information. Recently, RBM and DAE-based online classifier with an \textit{evolving} architecture is developed \cite{onlineRBM2016chen,onlineAE2012Zhou,ashfahani2020devdan,ramasamy2020ORBM}. The online RBM method discussed in \cite{ramasamy2020ORBM} considers only adding neurons and not pruning redundant neurons that are no longer significant and relevant. In addition, the data classification problems are addressed by adapting the network with hyperparameters thresholds to evolve the architecture. Evolving denoising autoencoder discussed in \cite{ashfahani2020devdan} evolves the network neurons on-demand, which is also capable of extracting the meaningful latent representations from the data streams but has the limitation of using labels during fine-tuning the weights and has been adapted for data classification in a mini-batch mode. Online clustering methods for data streams have been addressed in \cite{cormode2005s} and \cite{datar2002estimating}. Online data mining algorithms face a problem with the high dimensionality of input data (dimensional disaster or curse of dimensionality). To overcome this issue, dimensionality reduction algorithms like RBM (\cite{yuan2018clustering,Senthil2024rbm}) are used to transform the incoming stream data into a lower-dimensional feature space before grouping it into clusters. Using this as inspiration, the proposed method aims to transform the incoming high-dimensional data from data streams into a lower dimension before applying clustering approaches to cluster the data while at the same time evolving the RBM architecture to adapt to the corresponding data stream.

\begin{figure*}[t]
\centering
\includegraphics[width=0.8\textwidth]{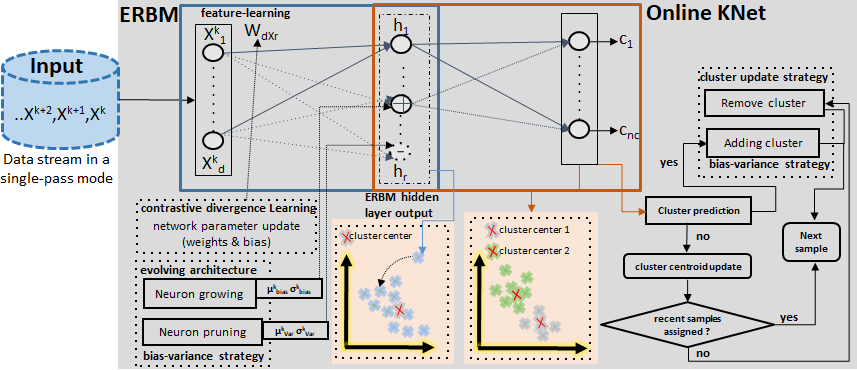} 
\caption{Schematic overview of the ERBM-KNet.}
\label{Fig1}
\vspace{-0.5cm}
\end{figure*}

\section{ERBM-KNet method}\label{ERBM_method}
The proposed ERBM-KNet integrates the key ideas of evolving architecture and online clustering (cluster prediction and cluster center update). The flow of the ERBM-KNet is shown in Figure \ref{Fig1}.

\noindent\textbf{Preliminaries:} Let us consider an online clustering task where unlabelled data (denoted by $X^k$) is arriving sample-by-sample at $k^{th}$ instance, $X^k \in \Re^{d}$. Here, $d$ is the number of input features. The goal is to map a high-dimension to low-dimensional latent space $f: \Re^{d}\rightarrow \Re^{r}$ to generate uncorrelated data distribution while minimising the reconstruction error. RBM is trained using the one-step contrastive divergence (CD-1) \cite{hinton2007boltzmann,Senthil2024rbm}, which is simplified by conditional probability ($P(z|x)$ and $P(x|z)$) in a CD-1 by maximizing data log-likelihood distribution. Thus a better representation is achieved so that data reconstruction ($\hat{X}$) is closer to the input $X$. Previously, ORBM \cite{onlineRBM2016chen,ramasamy2020ORBM} uses reconstruction error to add neurons. In this study, ERBM uses Network Significance (NS) as a condition to alter the network architecture similar to that used in mini-batch Autoencoder classifier \cite{ashfahani2020devdan} to add and remove the hidden neurons.

\subsection{ERBM architecture}
The concept of NS was used in \cite{ashfahani2020devdan} for evolving an Autoencoder for an online classifier in a mini-batch mode. Here, NS is used to determine the information extraction capabilities of the ERBM for online clustering in a single-pass mode. 

\noindent\textbf{Analysis of bias-variance in ERBM:}
NS is expressed as
\begin{equation}
NS = \int_{-\infty}^{\infty}(X-\hat{X})^{2}p(X)dX = E[(X-\hat{X})^{2}],
\label{NS}
\end{equation}
where $X$ is the input sample to the network, $\hat{X}$ is the reconstructed input obtained after passing through the RBM network and $p(X)$ is the probability density function of the input. Here, (\ref{NS}) can be simplified and expressed as
\begin{equation}
     NS = Bias(\hat{X})^{2} + Var(\hat{X}),
     \label{Rewritten equation}
\end{equation}
where the bias and variance terms are given by E($X$ - E[$\hat{X}$])$^{2}$ and (E[$\hat{X}^{2}$] - E[$\hat{X}$]$^{2}$), respectively. E[$\hat{X}$] is computed using
\begin{equation}
E[\hat{X}] = s(E[z]W' + c).
\label{Exhat}
\end{equation}
where $z$ is the extracted features for a given sample, $W$ is the weight matrix between the RBM layers, $c$ is the hidden layer bias vector and $s$ is the sigmoid function. E[z] is given by
\begin{gather}
E[z] = \int_{-\infty}^{\infty}s(A)p(A)dA,
\label{Ey}
\end{gather}
where $A = XW + b$, $b$ is the input layer bias in the RBM and
$\ P(A) = \frac{1}{\sqrt{2\pi(\sigma_A^2)}}e^{-\frac{(A-\mu_A)^2}{2\sigma_A^2}}$ 
 is the normal probability density of $A$ with mean $\mu_A$ and variance $\sigma_A$. With the probit approximation \cite{murphy2012machine}, (\ref{Ey}) can be expressed as:
\begin{gather}
E[z] \approx s(\frac{\mu_{A}}{\sqrt{1 + \pi \sigma_{A}^{2}/8 }}).
\label{HS}
\end{gather}
where $\mu_{A}$ and $\sigma_{A}$ are the mean and standard deviation of $A$ which are computed for every sample. Let us now substitute (\ref{HS}) in (\ref{Exhat}) to obtain the expectation over the reconstructed inputs ($E[\hat{X}]$). To evaluate the expectation of $E[\hat{X}^2]$, recall that the product of two independent identically drawn (i.i.d.) variables gives $z^2$=$z*z$ then
\begin{gather}
\nonumber E[\hat{X}^2] = s(E[z^2]W' + c) = s(E[z]*E[z]W' + c)\\
            E[\hat{X}^2] = s(E[z]^2W'+c). \vspace{-1em}
\label{Ex2hat}
\end{gather}
After substituting (\ref{HS}) into (\ref{Ex2hat}), $E[\hat{X}^2]$ is obtained. The two expectation values, ($E[\hat{X}]$ and $E[\hat{X}^2]$), can subsequently be used to compute the bias and variance by substituting in (\ref{Rewritten equation}) to obtain the NS value that is used to alter the network architecture. 

\noindent\textbf{ERBM online learning mechanism:} ERBM can alter its structure to grow or prune hidden neurons to accommodate the continuous data. The objective is to alleviate the problem of high bias (underfitting) and high variance (overfitting) in the network represented as follows:
\begin{equation}
     \mu^{k}_{bias} + \sigma^{k}_{bias} \geq \mu^{min}_{bias} + d_1\sigma^{min}_{bias}.
     \label{High Bias}
\end{equation}
\begin{equation}
     \mu^{k}_{Var} + \sigma^{k}_{Var} \geq \mu^{min}_{Var} + 2d_2\sigma^{min}_{Var}.
     \label{High Variance}
\end{equation}
where $\mu^{k}_{bias}$, $\sigma^{k}_{bias}$ and $\mu^{k}_{Var}$ and $\sigma^{k}_{Var}$ are the recursive mean and standard deviation of the bias and variance, respectively. $\mu^{min}_{bias}$, $\sigma^{min}_{bias}$ and $\mu^{min}_{Var}$ and $\sigma^{min}_{Var}$ are the minimum value of the mean and standard deviation of the bias and variance for all the samples encountered. Furthermore, $d_1$ is a dynamic constant calculated as $(\alpha e^{(-(bias(\hat{X}))^2)} + \beta)$, where $\alpha$ and $\beta$ are constants, and the bias is the value computed for the current data sample. The parameters $\mu^{min}_{bias}$ and $\sigma^{min}_{bias}$ are reset every time the condition in (\ref{High Bias}) is satisfied by adding a neuron. When a neuron is added, a vector proportional to the input sample is appended to the weight matrix with a random value. In the case of pruning strategy, $d_2$ is calculated as $(\gamma e^{(-(Var(\hat{X}))^2)} + \delta)$, $\gamma$ and $\delta$ are constants, and the variance is the value computed for the current data sample. The parameters $\mu^{min}_{Var}$ and $\sigma^{min}_{Var}$ are reset every time the condition in (\ref{High Variance}) is satisfied by pruning a neuron. Once the RBM structure is updated, learning step commences using Contrastive divergence (CD) \cite{hinton2002training}.

\noindent\textbf{Contrastive divergence learning.} Let us assume that the probability distribution over the input ($x$) and the latent representation ($z$) is given by
\begin{equation}
     p(x;z) = \frac{1}{Q(x;z)}e^{-E(x;z)},
     \label{ProbabilityDistribution}
\end{equation}
where $Q$ is the normalization constant given by $\sum_{x,z}e^{-E(x;z)}$. The probability that the RBM structure assigns to a given data sample $x$ can be obtained by summing over all possible hidden vectors.
\begin{equation}
     p(x) = \sum_{z}\frac{1}{Q(x;z)}e^{-E(x;z)},
     \label{Probabilityx}
\end{equation}

The probability $p(x)$ in  (\ref{Probabilityx}) can be increased by tuning the weights and the layer bias. The maximum-likelihood estimate of the weights and biases for a given data sample can be determined using the gradient-ascent algorithm. From \cite{hinton2012practical} we have
\begin{equation}
    W_{i+1}  =  W_{i} + \eta \frac{\partial log(p(x))}{\partial W}
    \label{W1x}
\end{equation}

\noindent The derivative of the log probability in (\ref{W1x}) is expressed as
\begin{equation}
     \frac{\partial log(p(x))}{\partial W} = <xz>_{data} - <xz>_{model}
    \label{derivativex}
\end{equation}
The angle brackets are used to denote expectations under the distribution specified by the subscript that follows. The first term in (\ref{derivativex}) can be computed by multiplying the probability values that are obtained after performing a forward-pass and backward-pass through the RBM structure for a given data sample. Computing the second term is significantly more challenging. In \cite{hinton2002training}, a technique called contrastive divergence was proposed wherein the input data distribution can be captured by replacing the second expectation term over the entire input data with a point estimate that is obtained by performing Gibbs sampling twice on the Markov Chain. Thus, in this paper, Contrastive Divergence-1 (CD-1) is employed to train the RBM for all input samples to ease the computational burden. This study adds a momentum term to achieve faster convergence and prevent convergence at local optima. The weight and bias are updated using CD-1 as
\begin{gather}
\nonumber W_{i+1}  =  vW_{i} + \eta*(c_{1} - c_{2})\\
vW_{i+1} = vW_{i} + W_{i+1}
\label{Weight Update}
\end{gather}
where $v$ is the momentum coefficient, $\eta$ is the learning rate, and $c_{1}$ and $c_{2}$ are defined by 
\begin{gather}
\nonumber c_{1} = (z_{0})^{T}X,\\
c_{2} = (z_{1})^{T}\hat{X}
\label{C Value}
\end{gather}
\noindent
where $z_{0}$ and $z_{1}$ are the output of the hidden layer for the original input and the reconstructed input respectively. The bias update are given by
\begin{gather}
\nonumber b_{i+1} = vb_{i} + \eta*(X - \hat{X}), \\
 vb_{i+1} = vb_{i} + b_{i+1}, \\ 
\nonumber c_{i+1} = vc_{i} + \eta*(z_{0} - z_{1}), \\
 vc_{i+1} = vc_{i} + c_{i+1}. 
\label{bias paramupdate}
\end{gather}
\noindent\textbf{Reconstruction error.} Conventional RBM adapts conditional probability in a CD with $P(z|x)$ and $P(x|z)$ by maximizing data log-likelihood, which helps in data reconstruction closer to the original data. Therefore, the likelihood distribution implicitly reflects the reconstruction error. In the literature, the Online RBM \cite{onlineRBM2016chen} uses reconstruction error to add neurons. In this study, we use NS as a condition to alter the network architecture similar to that used in evolving denoising autoencoder-based classifier \cite{ashfahani2020devdan} for adding and removing the hidden neurons.

\noindent\textbf{Hidden layer output assumption.} The hidden layer output is not assumed to be normally distributed. The derivation in (\ref{HS}) is a simple substitution of the "logit" function with a "probit" function. According to \cite{murphy2012machine}, the "logit" function can be approximated with a "probit" function ($s(x) \approx  \Phi(\lambda x) $) regardless of the input data distribution for suitable values of lambda ($\lambda^{2}$ is $\pi/8 $). This property is used to evaluate an expectation (\ref{HS}) that otherwise does not have an analytical solution as expressed in \cite{murphy2012machine}.

\subsection{KNet autonomous online clustering}
\textbf{Cluster prediction:} The proposed algorithm is capable of predicting the number of clusters by using a process similar to the NS approach used for growing and pruning the number of hidden neurons. Here, we use the L2 norm as the compatibility measure between an input data sample's latent representation and the cluster centers. 
The condition to add a new cluster is given by:
\begin{equation}
      \min_{i=0,\dots n_c} D(C_{i},HR_{1}) > \mu_{D} + d_3\sigma_{D},
     \label{Cluster Update}
\end{equation}
where $n_c$ is the number of clusters, $\mu_{D}$ and $\sigma_{D}$ are the mean and standard deviation of the distance between a latent sample ($HR_{1}$) and its cluster center ($C_{i}$), and $d_3$ is the dynamic constant is employed to realize a dynamic confidence degree is calculated as $(\eta e^{-minDist} + \nu)$, where $\eta$ and $\nu$ are constants, and "minDist" is the Euclidean distance or average Euclidean distance between the closest cluster center and the latent representation of the current data sample \cite{Andri2022Clust}.

A novel cluster removal strategy is proposed, let $x^k$ be the current sample at $k^{th}$ time step, and $Cluster\_added$ be a number to record the latest cluster added. Suppose $x^k$ comes in, we add a new cluster, then $Cluster\_added = k$ and $[s_1,s_2,...,s_n]$ is a list of numbers saving the latest time step when a sample is assigned to cluster $i$, $i\in \{1,...,n\}$. Then the condition to remove a cluster is given by:
\begin{align}
\label{cluster remove1}\textbf{Condition1:}& k-Cluster\_added \geq threshold_1,\\
\label{cluster remove2}\textbf{Condition2:}& k-s_i \geq threshold_2, 
\end{align}
\begin{equation}
\nonumber \textbf{IF} (Condition1) \text{AND} (Condition2) \textbf{THEN}\quad remove\quad cluster_i
\end{equation}
where $Condition1$ guarantee for a sufficient time that no new clusters are generated. Every sample comes to the old cluster centers, giving each center time to attract samples. The $Condition2$ means that even in this case, $cluster_i$ still cannot win any sample to its cluster. Then we remove this $cluster_i$ since it may not be suitable for the current data distribution. We mainly apply this strategy to large datasets; as more samples coming in to give more chance to generate clusters, the model would wrongly generate cluster centers sometime. The thresholds, $threshold_1$ and $threshold_2$, are usually set to $1\% \sim 2\%$ of the number of samples. 

Though it may seem that the number of clusters is dependent on the input data distribution. This is not necessarily the case. For instance, consider a data stream consisting of 3 classes (A,B,C). Initially if only a few samples of A arrive followed by a random distribution of all the samples of only B and C, before the next sample of A arrives, then it is possible that the cluster assigned to class A is erroneously pruned because the condition in (\ref{cluster remove2}) is satisfied. This does not pose an issue because the cluster will be added once again when the algorithm encounters samples from class ``A". 

\begin{figure}[h]
\begin{center}
\includegraphics[height=3cm]{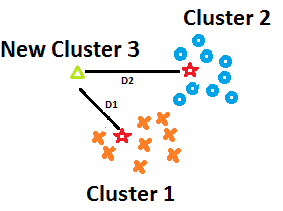}
\end{center}
\caption{New cluster addition}
\label{New Cluster Addition}
\end{figure}

Figure \ref{New Cluster Addition} demonstrates an instance of the cluster addition mechanism. The orange crosses and the blue circles represent two clusters in the hidden space (ERBM output space), and the red stars are their corresponding cluster centers. The green triangle is the hidden layer output for the current data sample. The distance from the current sample to two existing clusters are D1 and D2, respectively. From the Figure, since D1 $<$ D2, D1 is substituted on the left-hand side of (\ref{Cluster Update}). Assuming (\ref{Cluster Update}) is satisfied then a new cluster is initialized, and the hidden response for the current data sample (green triangle in Fig. \ref{New Cluster Addition}) is set as the cluster center for the new third cluster (with only 1 sample for the time being).

\begin{algorithm}
\fontsize{9}{10}\selectfont     \KwIn{clustering dataset \textbf{X}=${\{\textbf{x}^k\in {\rm I\!R}^d\}_{k=1}^N}$\\}
     \KwOut{$n_c$, NMI, Purity} 
     \par \While{\textbf{x}$^k$ in \textbf{X}}{
     \textbf{Calculate:} \text{ E[z] and E[$z^{2}$] using (\ref{HS}, \ref{Ex2hat}}) \par
     \textbf{Calculate:} \text{$\mu^{k}_{bias}, \sigma^{k}_{bias},\mu^{k}_{Var}, \sigma^{k}_{Var}$ using (\ref{Exhat}, \ref{Ex2hat})} \par
     \textbf{Neuron Growth:} \par

     \If{($\mu^{k}_{bias} + \sigma^{k}_{bias} \geq \mu^{min}_{bias} + k\sigma^{min}_{bias}$)}
     {
     \textbf{Initialization:} $W_{\text{new}} \gets c'X$, $c_{\text{new}} \gets 0$ \par
    \textbf{Reset:} $\mu^{min}_{bias}$, $\sigma^{min}_{bias}$,
    $N \gets N + 1$
    GFlag $\gets$ 1
    }

    \Else
    {GFlag $\gets$ 0} \par

    \textbf{Neuron Pruning:} \par
    
    \If {($\mu^{t}_{Var} + \sigma^{t}_{Var} \geq \mu^{min}_{Var} + 2k\sigma^{min}_{Var}$) \text{and} (GFlag != 1) \text{and} (N $>$ 1)} 
    {\text{Prune neuron with Least E[z]} \par
    \textbf{Reset:} $\mu^{min}_{Var},\sigma^{min}_{Var}$, N $\gets$ N - 1, PFlag $\gets$ 1 
    } 

    \Else
    {PFlag $\gets$ 0}
    \text{Update ERBM parameters using CD ([\ref{Weight Update}] -[\ref{bias paramupdate}])} \par
    \text{Determine the number of clusters using (\ref{Cluster Update}}) \par
    {Update cluster centres using (\ref{SOM weight Update}-\ref{HR update SOM})} \par
    {Remove cluster centres using (\ref{cluster remove1}-\ref{cluster remove2})}
     }
     \caption{ERBM-KNet}
\end{algorithm}
\noindent\textbf{Online clustering.} Online KNet-based data clustering is applied to the ERBM-extracted features (latent representation). KNet is a two-layer NN architecture (input layer maps directly with the output layer). For data clustering, $K$ Gaussian kernels are randomly initialized, and a standard deviation of $\sigma$. The standard deviation allows the neighbouring neurons to be updated along with the winning neuron. The learning starts with cooperation between all the neurons, i.e., for a given data sample, all the centers adjust their location. As the training progresses and the standard deviation of the Gaussian corresponding to each center begins to decrease with the accumulation of continuous data. This results in a reduced region of influence for the centres, which corresponds to the adaptation phase of the competitive learning framework. The update equation for a given neuron $c$ is given by:
\begin{equation}
    w_{c} = w_{c} + \eta h_{ic}(x_{i} - w_{c}),
    \label{SOM weight Update}
\end{equation}
where $\eta$ is the learning rate that decays with the number of samples and $h_{ic}$ is the neighbourhood kernel function given by 
\begin{equation}
    h_{ic} =\exp({\frac{-d_{ic}^{2}}{2\sigma_{c}^{2}}}),
    \label{HR update SOM}
\end{equation}
where $d_{ic}$ is the Euclidean distance between $w_{i}$ and $w_{c}$. The proposed ERBM-KNet is summarized in \textbf{Algorithm 1}\footnote{Code availability: Code for this work will be made available upon acceptance of the paper, in compliance with our organizational policies.}.
\begin{table}[t]
 \centering
 \caption{Summary of benchmark datasets.}
\begin{tabular}{cccccc}
\hline
Dataset & \#Dimensions & \#Training Samples  & \#Classes\\
\hline
MNIST & 28x28 & 60,000  &  10\\
Fashion &  28x28 & 60,000 & 10\\
Kuzushiji & 28x28 & 60,000 & 10\\
Coil-20 & 32x32   & 1,440  & 20  \\
Wafer & 52x52 & 4,015 & 5 \\
\hline
\end{tabular}
\label{Datasets}
\end{table}

\begin{table}[h]
 \centering
 \caption{Number of neurons evolved by ERBM.}
\begin{tabular}{cc}
\hline
Datasets & \#neurons \\
\hline
MNIST & 78.57$\pm$16.30 \\
Fashion & 39.71$\pm$8.80\\
Kuzushiji & 91.71$\pm$6.15\\
Coil-20 & 24.57$\pm$0.72 \\
Wafer & 40.14$\pm$0.98 \\
\hline
\end{tabular}
\label{ERBM_neurons}
\end{table}

\noindent\textbf{Time complexity.} The complexity of the ERBM algorithm with the CD method is largely due to the naive matrix production. For example, the time complexity of a naive matrix production for matrix A with size (\textit{a}*\textit{b}) and matrix B with size (\textit{b}*\textit{c}) is $O(a*b*c)$. Furthermore, clustering is applied to the reduced latent dimension. The KNet algorithm time complexity is $O(n_h*n_c)$ where $n_h$ is the number of hidden neurons evolved in ERBM, and $n_c$ is the number of current cluster centers. Hence, the overall time complexity of ERBM-KNet is $O(N*(1*n_v*n_h + n_h*n_c))$, where $N$ is the number of samples, and $n_v$ is the number of visible units, which is the dimension of raw input data and 1 is for the single-pass mode.

\begin{table*}[t]
  \centering
      \caption{Performance comparison of ERBM-KNet against other offline and online clustering methods (in percentage)}
  \fontsize{9}{9}\selectfont
   \begin{tabularx}{\textwidth}{p{8.2em}|p{4em}|p{7em}|p{7em}|p{7em}|p{7em}|p{7em}}
    \hline
    \textbf{Approach} & \textbf{Metric} & \textbf{MNIST} & \textbf{FMNIST} & \textbf{KMNIST} & \textbf{Coil-20} & \textbf{Wafer} \\
    \hline 
    \textbf{PCA-Kmeans} \cite{xu2015pca} & \begin{tabular}[c]{@{}l@{}}NMI\\ Purity\end{tabular} & 
    
   \begin{tabular}[c]{@{}r@{}}$46.51\pm0.26$\\ $57.48\pm0.38$\\ \end{tabular}    
    
    &  
    
   \begin{tabular}[c]{@{}r@{}}$51.12\pm0.97$\\ $\textbf{55.94}\pm1.07$\\ \end{tabular}
   
   &  
    \begin{tabular}[c]{@{}r@{}}$46.23\pm0.08$\\ $59.32\pm0.06$\\ \end{tabular}

    &     \begin{tabular}[c]{@{}r@{}} $37.84\pm1.77$ \\ $14.24\pm1.04$ \\ \end{tabular}            
    
    &         
    
    \begin{tabular}[c]{@{}r@{}} $75.13\pm0.89$ \\ $68.86\pm1.27$ \\ \end{tabular}

     \\ [1 em] 
     \vspace{-1em}
     \textbf{PCA-FCM} \cite{dogruparmak2014using} \vspace{-1em} & \vspace{-1.5em}\begin{tabular}[c]{@{}l@{}}NMI\\ Purity\end{tabular} \vspace{-1em}& 
    
    \vspace{-1.5em}\begin{tabular}[c]{@{}r@{}}$47.70\pm1.48$\\ $58.68\pm1.63$ \end{tabular}            
    \vspace{-1em}& 
    
    \vspace{-1.5em}\begin{tabular}[c]{@{}r@{}}$51.54\pm0.62$\\ $56.84\pm1.12$ \end{tabular}             
\vspace{-1em}&  
\vspace{-1.5em}
\begin{tabular}[c]{@{}r@{}}$44.14\pm0.05$\\ $57.16\pm0.04$\end{tabular}

\vspace{-1em}&  \vspace{-1.5em}  \begin{tabular}[c]{@{}r@{}} $38.40\pm1.13$ \\ $15.0\pm0.83$ \end{tabular}

\vspace{-1em}&  \vspace{-1.5em}  \begin{tabular}[c]{@{}r@{}}
$75.68\pm0.39$\\ $68.62\pm1.47$ \end{tabular}

\\ \vspace{-1em}

    \textbf{PCA-SOM} \cite{lopez2004principal} & \vspace{-1.5em}\begin{tabular}[c]{@{}l@{}}NMI\\ Purity\end{tabular} \vspace{-1em} & \vspace{-1.5em}

    \begin{tabular}[c]{@{}r@{}}$46.89\pm1.05$\\ $58.51\pm0.55$\end{tabular}           \vspace{-1em}   &

 \vspace{-1.5em}  \begin{tabular}[c]{@{}r@{}}$51.80\pm0.85$\\ $58.68\pm2.31$\end{tabular}
   \vspace{-1em}
   &  
    
   \vspace{-1.5em}\begin{tabular}[c]{@{}r@{}}$44.01\pm0.56$\\ $57.34\pm0.24$\end{tabular}
    
    \vspace{-1em}
    & \vspace{-1.5em}  \begin{tabular}[c]{@{}r@{}} $40.92\pm2.46$ \\ $15.13\pm1.43$\end{tabular} 

    \vspace{-1em}
    & \vspace{-1.5em}  \begin{tabular}[c]{@{}r@{}} $73.79\pm0.42$ \\ $68.16\pm1.75$ \end{tabular}
    
    \\[1 em]

     \vspace{-1em}
    \textbf{SRBM-Kmeans} \cite{graphRBM} & \vspace{-1.5em}\begin{tabular}[c]{@{}l@{}}NMI\\ Purity\end{tabular} &
    \vspace{-1.5em}
  \begin{tabular}[c]{@{}r@{}} $51.94\pm1.13$ \\ $61.17\pm1.09$ \end{tabular}              &    
    \vspace{-1.5em}
    \begin{tabular}[c]{@{}r@{}} $49.68\pm2.51$ \\ $51.90\pm3.48$ \end{tabular}
    
    & 
\vspace{-1.5em}
    \begin{tabular}[c]{@{}r@{}} $45.33\pm1.45$ \\ $57.87\pm1.73$ \end{tabular}
    \vspace{-1em}
    & \vspace{-1.5em} \begin{tabular}[c]{@{}r@{}} $34.24\pm1.48$ \\ $14.48\pm2.34$\end{tabular}               
    \vspace{-1em}
    & \vspace{-1.5em}  \begin{tabular}[c]{@{}r@{}} $82.15\pm0.58$ \\ $71.38\pm1.32$\end{tabular}  
    
    \\ [1 em]
\vspace{-1em}
    \textbf{SRBM-FCM} \cite{yang2015deep} &\vspace{-1.5em} \begin{tabular}[c]{@{}l@{}}NMI\\ Purity\end{tabular} \vspace{-1em} & 
      
     \vspace{-1.5em} \begin{tabular}[c]{@{}r@{}}$51.46\pm0.84$\\ $61.02\pm0.93$\end{tabular}
      \vspace{-1em}
      & \vspace{-1.5em} \begin{tabular}[c]{@{}r@{}}$49.22\pm1.71$\\ $50.41\pm2.54$ \end{tabular}
      
      \vspace{-1em}&  \vspace{-1.5em}
      \begin{tabular}[c]{@{}r@{}}$46.28\pm0.61$\\ $58.92\pm0.82$\end{tabular}
      
      & \vspace{-1.5em} \begin{tabular}[c]{@{}r@{}} $33.78\pm1.90$ \\ $15.38\pm2.6$\end{tabular}              
      
      \vspace{-1em}
      &\vspace{-1.5em} \begin{tabular}[c]{@{}r@{}} $81.48\pm2.63$ \\ $71.69\pm1.08$\end{tabular}
      
      \\ [1 em]
    \vspace{-1em}
     \textbf{SRBM-SOM} \cite{Senthil2024rbm} & \vspace{-1.5em}\begin{tabular}[c]{@{}l@{}}NMI\\ Purity\end{tabular} & \vspace{-1.5em}
    \begin{tabular}[c]{@{}r@{}} $51.05\pm0.74$\\ $61.82\pm1.48$\end{tabular}
    \vspace{-1em}&  \vspace{-1.5em}
    
    \begin{tabular}[c]{@{}r@{}}$50.88\pm2.40$\\ $53.10\pm3.20$\end{tabular}   
    
   \vspace{-1em} &  \vspace{-1.5em}
    
    \begin{tabular}[c]{@{}r@{}}$44.26\pm2.04$\\ $56.17\pm3.07$\end{tabular}
\vspace{-1em}
    & \vspace{-1.5em} \begin{tabular}[c]{@{}r@{}} $34.67\pm0.46$ \\ $15.53\pm2.53$\end{tabular}                
    \vspace{-1em}
    & \vspace{-1.5em}\begin{tabular}[c]{@{}r@{}} $\textbf{82.20}\pm1.35$ \\ $71.38\pm2.51$ \end{tabular}
    
    \\[-1em] \hline
\vspace{-1em}
       \textbf{OFCM} \cite{Hall2011OnlineFCM} & \begin{tabular}[c]{@{}l@{}}NMI\\ Purity\end{tabular} & \begin{tabular}[c]{@{}r@{}}  $25.23\pm0.0$\\ $36.74\pm0.0$\end{tabular}
       \vspace{-1em}
       &   \begin{tabular}[c]{@{}r@{}} $42.90\pm0.0$ \\ $47.37\pm0.0$ \end{tabular}             
       \vspace{-1em}
       &  \begin{tabular}[c]{@{}r@{}} $16.19\pm0.0$ \\ $26.13\pm0.0$ \end{tabular}              
    \vspace{-1em}
    &    \begin{tabular}[c]{@{}r@{}} $58.13\pm0.0$ \\ $39.57\pm0.0$\end{tabular}
    \vspace{-1em}
    &\begin{tabular}[c]{@{}r@{}} $54.25\pm0.0$ \\ $48.16\pm0.0$ \end{tabular}                
    
    \\ [1em]
\vspace{-1em}
     \textbf{OKmeans} \cite{Vincent2021OKmeans} & \vspace{-1.5em}\begin{tabular}[c]{@{}l@{}}NMI\\ Purity\end{tabular} & \vspace{-1.5em} \begin{tabular}[c]{@{}r@{}} $38.70\pm0.0$ \\ $50.45\pm0.0$ \end{tabular}
     \vspace{-1em}
     & \vspace{-1.5em}   \begin{tabular}[c]{@{}r@{}} $52.46\pm0.0$\\ $52.97\pm0.0$\end{tabular}     
    \vspace{-1em}
    &  \vspace{-1.5em} \begin{tabular}[c]{@{}r@{}} $41.42\pm0.0$\\ $51.96\pm0.0$ \end{tabular}
    \vspace{-1em}
    &  \vspace{-1.5em}   \begin{tabular}[c]{@{}r@{}} $\textbf{64.37}\pm0.0$ \\ $42.35\pm0.0$ \end{tabular}             
    \vspace{-1em}
    &  \vspace{-1.5em}  \begin{tabular}[c]{@{}r@{}} $41.17\pm0.0$ \\ $42.36\pm0.0$ \end{tabular} 
    \\ [0.3em] \hline
\vspace{-1em}

    \textbf{ERBM-KNet} & \begin{tabular}[c]{@{}l@{}}NMI\\ Purity\\ Clusters\end{tabular}
    \vspace{-1em}& 
    
    \begin{tabular}[c]{@{}r@{}} $\textbf{52.93} \pm 2.11$ \\ $\textbf{63.19} \pm 4.14$ \\ $\textbf{10}$ \end{tabular} 
    \vspace{-1em}
    & \begin{tabular}[c]{@{}r@{}} $\textbf{54.63}\pm2.42$\\ $55.90\pm4.00$\\ $11$ \end{tabular}  

    \vspace{-1em}
    &  
    \begin{tabular}[c]{@{}r@{}} $\textbf{46.45}\pm1.18$ \\ $\textbf{60.66}\pm1.94$ \\ $\textbf{10}$ \end{tabular} 
    
    \vspace{-1em}
    &  \begin{tabular}[c]{@{}r@{}} $61.83\pm2.01$ \\ $\textbf{47.54}\pm1.47$ \\ $\textbf{20}$ \end{tabular}
    \vspace{-1em}
    &   \begin{tabular}[c]{@{}r@{}} $80.60\pm0.54$ \\ $\textbf{90.6}\pm1.14$ \\ $\textbf{5}$   \end{tabular}
    \\\hline
  \end{tabularx}
     \label{tab:table3}
  \vspace{-1.0em}
\end{table*}

\section{Experiments}
Comprehensive experiments have been conducted to evaluate the performance of the proposed ERBM-KNet in comparison to eight state-of-the-art methods across five image datasets. 

\subsection{Experimental settings}
\noindent\textbf{Datasets:} The image datasets (Table \ref{Datasets}) used in this study are MNIST \cite{LCun1998MNIST}, Fashion MNIST \cite{xiao2017fashion}, KMNIST \cite{Tarin2018KMNIST}, Coil-20 \cite{nene1996columbia}, and a semiconductor industry wafer defect dataset \cite{wang2020deformable}. The MNIST, Fashion MNIST, and KMNIST datasets consist of 10 classes each, whereas the COIL-20 and Wafer datasets contain 20 and 5 classes, respectively.

\noindent\textbf{Comparison with state-of-the-art methods:} The proposed ERBM-KNet was compared with two categories of state-of-the-art methods, i) Six feature-learning-based offline clustering methods (Figure~\ref{fig:user_intention}(b)): PCA-Kmeans \cite{xu2015pca}, PCA-FCM \cite{dogruparmak2014using}, PCA-SOM (self organizing map) \cite{lopez2004principal}, and Single-layer RBM (SRBM) methods: SRBM-Kmeans \cite{graphRBM}, SRBM-FCM \cite{yang2015deep} and SRBM-SOM \cite{lan2017deep,Senthil2024rbm}.
In SRBM-based feature-learning, the dataset is processed in a batch mode with the dimension same as the number of hidden neurons generated by the ERBM to make a fair comparison.
ii) Two baseline online clustering methods (Figure~\ref{fig:user_intention}(c)), namely, online Kmeans (OKmeans) \cite{Vincent2021OKmeans} and online FCM (OFCM) \cite{Hall2011OnlineFCM}. 

\noindent\textbf{Implementation details:}
The dynamic constant $d_1$ in (\ref{High Bias}) and $d_2$ in (\ref{High Variance}) are expressed as $(\alpha e^{-bias} + \beta)$ and $(\gamma e^{-Var^{2}} + \delta)$, respectively, where $\alpha,\beta,\gamma,\delta$ are hyperparameters that need to be defined. The values of the parameters $\alpha,\beta$ are in the interval [0.75,0.97] and [0.42,0.81] respectively. The values vary slightly for each dataset, the average value being 0.86$\pm$0.08 and 0.56$\pm$0.15. The parameters $\gamma$ and $\delta$ are in the interval [1,1.5] and [0.5,1] respectively. The $\alpha$ and $\beta$ limit the number of hidden neurons in the network. By increasing the values of $\alpha$ and $\beta$, the value of the dynamic threshold $d_1$ increases, and as a result, it increases the right-hand side of (\ref{High Bias}), thus making it harder for a neuron to be added to the ERBM hidden layer. The dynamic constant $d_3$ in (\ref{Cluster Update}) is defined similarly to that of the dynamic constant $d_1$, $(\eta e^{-minDist} + \nu)$. The values of the parameters $\eta$ and $\nu$ are in the interval [0.9,1.6] and [0.7,1.0] respectively. These two parameters are responsible for growing/limiting the number of clusters. By increasing the value of $\eta$ and $\nu$, the right-hand side of (\ref{Cluster Update}) increases, thus making it harder to add a new cluster. The number of hidden neurons obtained in ERBM after five runs on all the datasets are shown in Table \ref{ERBM_neurons}.

\noindent\textbf{Criteria for comparison:} When comparing cluster prediction, NMI (Normalized Mutual Information), and purity score, cluster prediction is considered the base criterion to determine whether a method has performed well. Therefore, we focus on tuning the hyperparameters (Table \ref{ERBM_neurons}) as discussed above to minimize the number of hidden neurons in the ERBM architecture, while still achieving a high-quality clustering performance. Finally, the metrics used for evaluating the performance of the algorithm are \textbf{purity} and \textbf{NMI}, both of which take value in the interval [0,100]. The parameters are calculated using equations (\ref{purity_equation}) and (\ref{NMI_equation}) from \cite{PutityNMIequation}, where \textit{n} in equation (\ref{purity_equation}) represents the number of samples and B = ($B_{1},B_{2},B_{3},...B_{k}$), A = ($A_{1},A_{2},A_{3},...A_{r}$) are sets of clusters obtained by a clustering method and ground truth respectively. I(X;Y) in (\ref{NMI_equation}) represents the mutual information between two discrete random variables X and Y, and H(X) represents the entropy of a discrete random variable X. The results are tabulated in Table \ref{tab:table3} in percentages as in \cite{graphRBM} with average and standard deviation for 5 seeds. 

\begin{equation}
    Purity(B,A) = \frac{1}{n} \sum_{k} \max_{r} |B_{k} \cap A_{r}|
     \label{purity_equation}
\end{equation}

\begin{equation}
    NMI(X,Y) = \frac{2I(X;Y)}{H(X)+H(Y)}
     \label{NMI_equation}
\end{equation}

\subsection{Comparison of clustering performance}
Table \ref{tab:table3} presents the performance comparison of several clustering algorithms on five different datasets. The evaluation metrics used to assess the performance of algorithms are NMI, Purity, and cluster prediction. The first set of algorithms evaluated in the table performs clustering in an offline manner using a fixed architecture. These algorithms typically employ PCA or SRBM as feature extraction methods, followed by Kmeans, FCM, or SOM clustering. The second set of algorithms performs clustering in an online manner (OFCM and OKmeans); however, there is no offline or online feature-learning embedded in this case. Finally, we present our approach  that combines online clustering with an evolving architecture.

Our proposed approach, ERBM-KNet, exhibits superior performance on the MNIST dataset, achieving the highest NMI and purity scores of 52.93 and 63.19, respectively. Furthermore, ERBM-KNet also accurately categorizes the dataset into 10 clusters. While the first set of algorithms also shows competitive performance, online clustering approaches without feature-learning are not effective for the MNIST dataset. On the Fashion MNIST dataset, ERBM-KNet outperforms other methods in terms of NMI score. ERBM-KNet achieves an NMI score of 54.63 with 11 clusters, while PCA-Kmeans achieves the highest purity score. In the case of the KMNIST dataset, our approach surpasses the performance of other methods and obtains the highest NMI and purity scores of 46.45 and 60.66, respectively, while PCA-Kmeans achieves a similar NMI score of 46.23. Notably, ERBM-KNet  successfully categorize the KMNIST dataset into the correct number of clusters. On the Coil-20 dataset, ERBM-KNet achieves a comparable NMI score of 61.83 and outperforms other methods in terms of the Purity metric with a score of 47.54. Conversely, the first set of algorithms perform poorly on the Coil-20 dataset, with a maximum NMI of 40.92. Finally, on the Wafer dataset, ERBM-KNet  stands out by achieving the highest Purity score of 90.6. Notably, ERBM-KNet accurately categorizes the Wafer dataset into the correct number of clusters, showcasing its ability to effectively handle complex semiconductor industry datasets with precise cluster assignments.

Remarkably, ERBM-KNet performs well across all datasets consistently, underscoring its robustness and effectiveness in various data settings. Moreover, ERBM-KNet shows its ability to automatically determine the appropriate number of clusters, highlighting its capacity to capture the intrinsic structure of the datasets. Importantly, these impressive results are achieved with an average number of 55 neurons across 5 datasets, further demonstrating the efficiency and scalability of ERBM-KNet. Overall, the proposed algorithm outperforms other baseline methods and feature-learning approaches, such as PCA and SRBM, in terms of clustering performance. Also, compared to online clustering algorithms for image datasets, our method consistently achieves higher purity, NMI score, and accurate cluster predictions.

\subsection{Reconstruction performance}
\begin{figure}[h]
\centering
\includegraphics[height=4.9 cm, width = 8.5 cm]{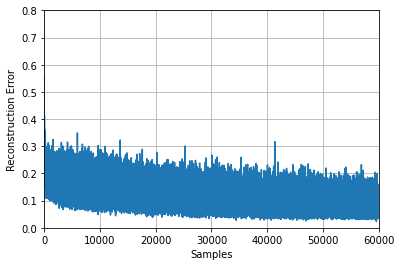}
\caption{ERBM reconstruction error for MNIST dataset.}
\label{reconerbm}
\end{figure}

The reconstruction performance of the proposed algorithm is compared with the ORBM architecture proposed in \cite{ramasamy2020ORBM} for the MNIST dataset. After training on half of the MNIST dataset, the ORBM algorithm requires more than 400 hidden neurons to capture the input data distribution effectively (for more details, please refer to Figure 3 in \cite{ramasamy2020ORBM}). In comparison, the proposed ERBM method requires an average of only 78.57 neurons to capture the data distribution for the entire dataset. From Figure \ref{reconerbm}, it is evident that the proposed ERBM method achieves a similar or lower reconstruction error (L2 Norm) when compared to the ORBM algorithm, with an 81.5\% reduction in the number of hidden neurons. 

\begin{figure}[]
\centering
\includegraphics[height=4.2 cm, width = 8 cm]{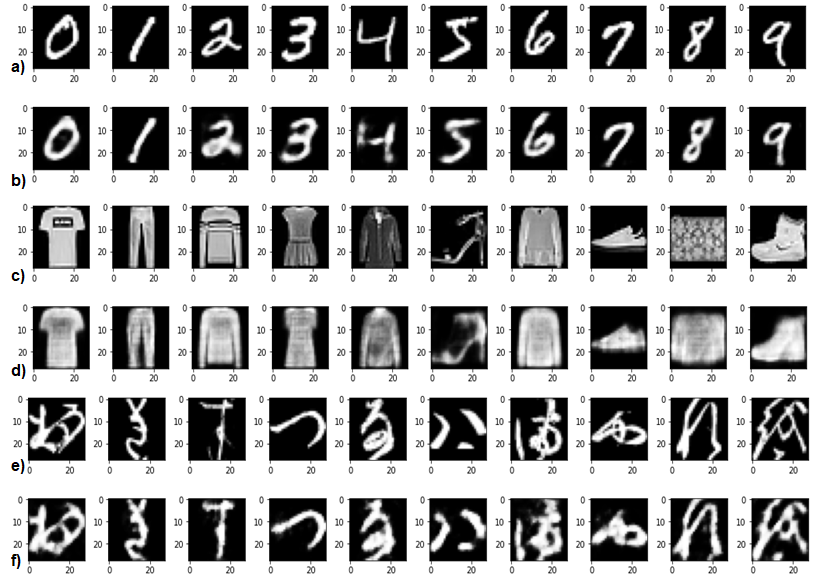}
\caption{original images a) MNIST, c) FMNIST, e) KMNIST; ERBM reconstructed images b) MNIST, d) FMNIST, f) KMNIST}
\label{erbm mnist reconstruction}
\end{figure}

\begin{figure}[]
\centering
\includegraphics[width = \linewidth]{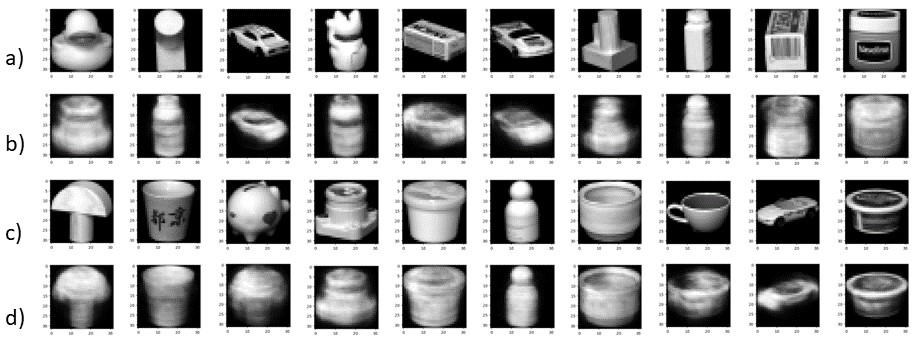}
\caption{a) and c) original Coil-20 images; b) and d) ERBM reconstructed images}
\label{erbm coil reconstruction}
\end{figure}

\begin{figure}[]
\centering
\includegraphics[width = \linewidth]{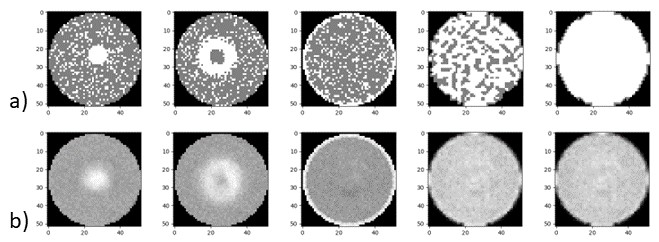}
\caption{a) original Wafer images; b) ERBM reconstructed images}
\label{erbm wafer reconstruction}
\vspace{-0.5cm}
\end{figure}

\noindent\textbf{ERBM reconstructed images:} Figure \ref{erbm mnist reconstruction} displays the reconstruction performance of ERBM with 74 hidden neurons for MNIST dataset. Figure \ref{erbm mnist reconstruction} displays the reconstruction performance of ERBM for MNIST, FMNIST and KMNIST. The images depicted in Figures \ref{erbm coil reconstruction} and \ref{erbm wafer reconstruction} showcase the successful reconstruction of the Coil-20 and Wafer datasets, respectively. The reconstructions utilize the power of ERBM, with a mere average of 25 neurons for Coil-20 and 40 neurons for the Wafer dataset. 

\begin{figure}[]
\centering
\includegraphics[height=4.5 cm, width = 8 cm]{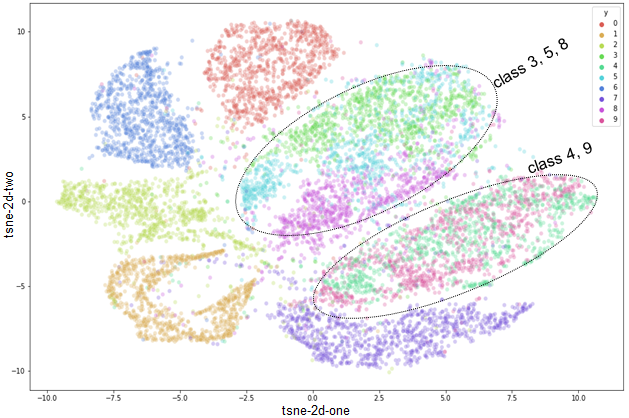}
\caption{Original t-SNE plot for MNIST dataset.}
\label{org mnist tsne}
\end{figure}

\begin{figure}[]
\centering
\includegraphics[height=4.5 cm, width = 8 cm]{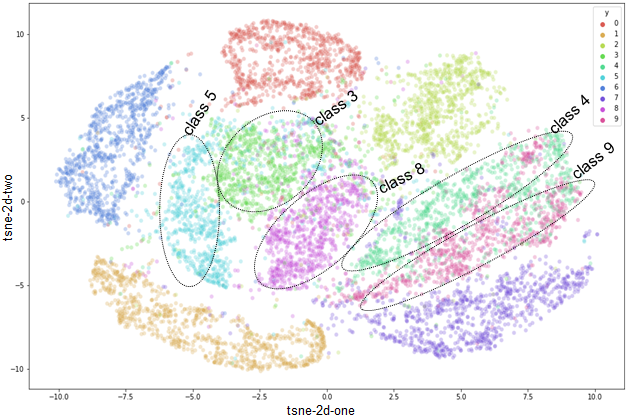}
\caption{ERBM modified t-SNE plot for MNIST dataset.}
\label{erbm mnist tsne}
\end{figure}

\noindent\textbf{Feature Visualization:} Figure \ref{org mnist tsne} represents the t-SNE (t-Distributed Stochastic Neighbor Embedding) of the original MNIST data distribution (784-dimensional), while Figure \ref{erbm mnist tsne} represents the data distribution in the 74-dimensional latent space generated by the ERBM. Interestingly, the nonlinear mapping of the data into the lower dimensional space has allowed the different classes to separate from one another. This is especially evident in the case of classes 4 and 9. In the original distribution (Figure \ref{org mnist tsne}), the two classes are heavily interspersed or overlapped and are not linearly separable. However, in Figure \ref{erbm mnist tsne} the two classes have been separated from one another to a great extent. The same is the case for classes 3, 5, and 8. The three classes are initially interspersed and mixed, but ERBM separates the classes and creates three distinct clusters. 

\begin{figure}[]
\centering
\includegraphics[height=4.5 cm, width = 8 cm]{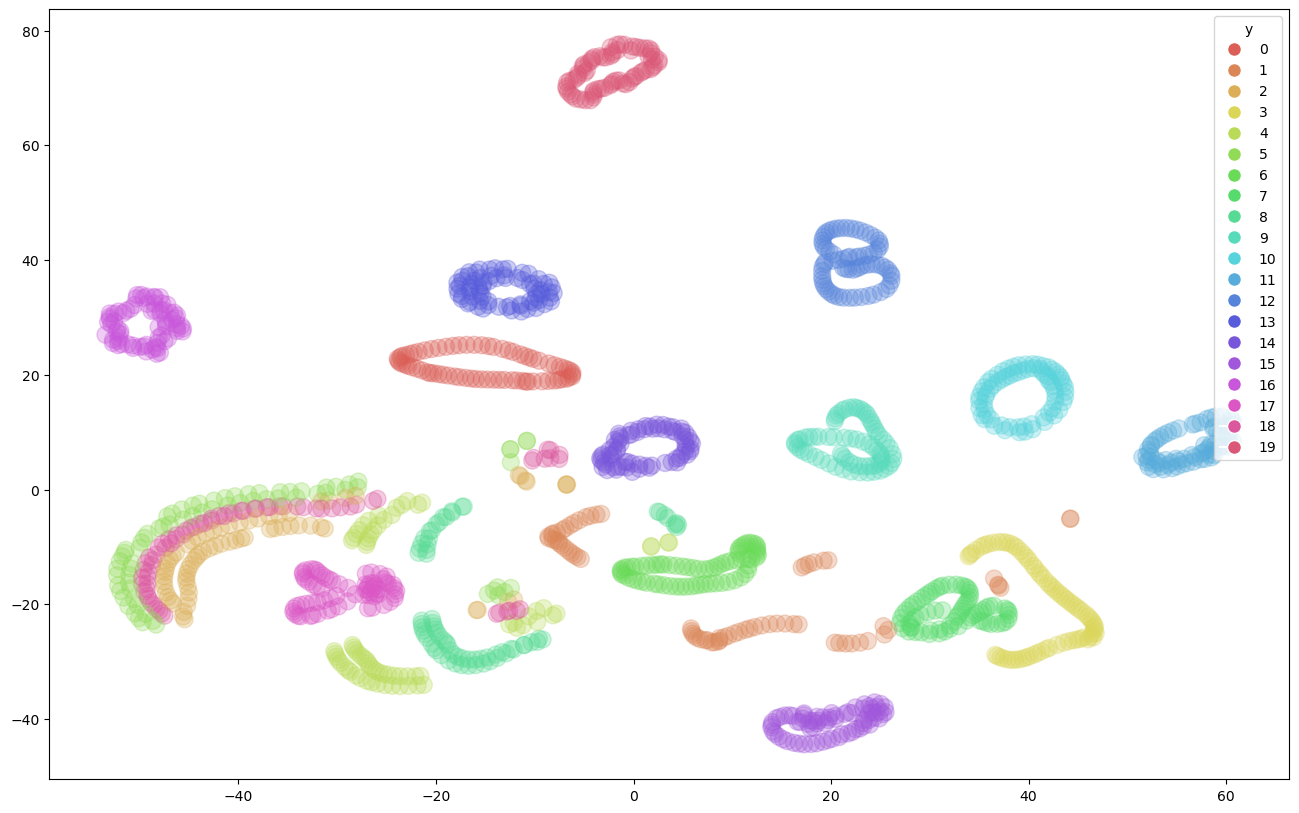}
\caption{Original t-SNE plot for Coil-20 dataset.}
\label{fig:rawtsnecoil}
\end{figure}

\begin{figure}[]
\centering
\includegraphics[height=4.5 cm, width = 8 cm]{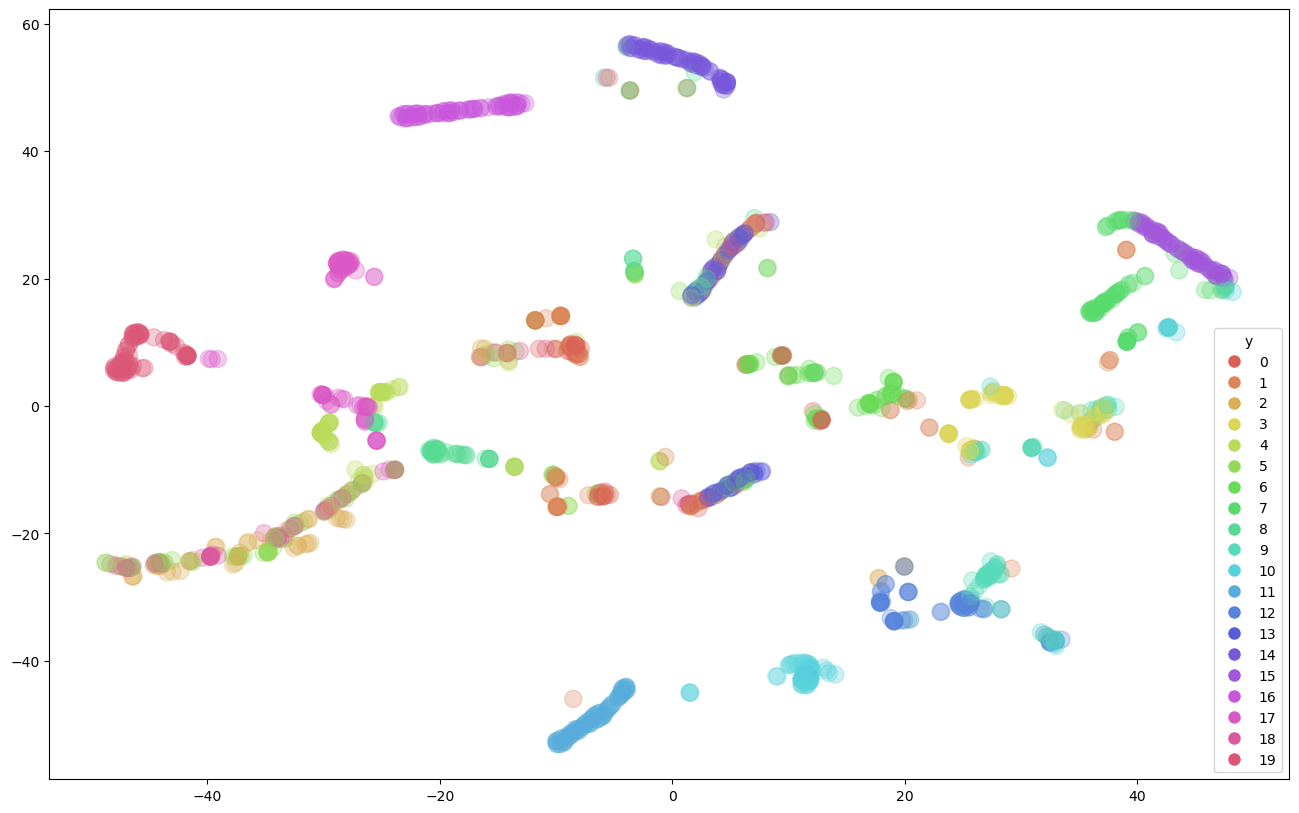}
\caption{ERBM modified t-SNE plot for Coil-20 dataset.}
\label{fig:erbmtsnecoil}
\end{figure}

Figure \ref{fig:rawtsnecoil} shows the t-SNE plot for the original data distribution where 11 clusters, i.e., \{1,10,11,12,13,14,15,16,17,18,20\}, out of 20 have high inter-cluster separation and low intra-cluster separation, which is easier to cluster the data. Whereas Figure \ref{fig:erbmtsnecoil} shows the representation of the dataset using ERBM where 6 clusters \{12,15,16,17,18,20\} out of 20 have high inter-cluster separation and low intra-cluster separation but other clusters get overlapped due to linearly separable data projected to the non-linear transformation. Hence for this dataset, OKmeans performs better than all the other methods, as shown in Table \ref{tab:table3}. 

\subsection{Ablation Study}
This ablation study summarizes the impact of hyperparameter variations, including $\alpha$, $\beta$, $\gamma$, and $\delta$, on the performance metrics of our model using the MNIST dataset, with each parameter varied one at a time while holding the others constant. Increasing $\alpha$ from 0.72 to 0.78 initially boosted NMI and Purity, suggesting improved cluster quality. However, higher values led to over-partitioning and decreased NMI and purity, indicating sensitivity to excessive $\alpha$. Similarly,  increasing $\beta$ from 0.38 to 0.42 enhanced NMI, Purity, and the number of clusters, implying better alignment with ground truth. But exceeding 0.42 reversed this trend, decreasing NMI and Purity. Increasing $\gamma$ from 1.0 to 1.2 also initially improved NMI and Purity, but further increases reduced both metrics and the number of clusters, eventually reaching the optimal number of clusters. $\delta$ was changed from 0.5 to 1.1, and in contrast, it had a consistently negative impact on NMI and Purity, indicating worsening alignment with ground truth.


\section{Conclusions}
In this paper, an evolving RBM (ERBM) for autonomous online clustering is proposed. The proposed ERBM-KNet can autonomously adapt the RBM architecture by growing and pruning the hidden neurons, using a bias-variance strategy, and also predicting the number of clusters in a data stream in an online fashion. Moreover, a detailed theoretical background is presented, including the derivation of the Network Significance criteria, which guides the evolution of the architecture. Finally, extensive experiments are conducted on four well-known benchmark datasets and an industry use case (wafer defect dataset) to test the proposed algorithm's performance in terms of both reconstruction and clustering. The ERBM-KNet is compared with feature-learning and baseline methods. The proposed ERBM-KNet significantly outperforms the state-of-the-art counterparts, reaffirming its effectiveness and practicality in online clustering scenarios. In our future research, robust deep ERBM architecture for online clustering will be explored.



\bibliographystyle{IEEEtran}
\bibliography{ERBMbib}

\begin{thebibliography}{10}
\providecommand{\url}[1]{#1}
\csname url@samestyle\endcsname
\providecommand{\newblock}{\relax}
\providecommand{\bibinfo}[2]{#2}
\providecommand{\BIBentrySTDinterwordspacing}{\spaceskip=0pt\relax}
\providecommand{\BIBentryALTinterwordstretchfactor}{4}
\providecommand{\BIBentryALTinterwordspacing}{\spaceskip=\fontdimen2\font plus
\BIBentryALTinterwordstretchfactor\fontdimen3\font minus \fontdimen4\font\relax}
\providecommand{\BIBforeignlanguage}[2]{{%
\expandafter\ifx\csname l@#1\endcsname\relax
\typeout{** WARNING: IEEEtran.bst: No hyphenation pattern has been}%
\typeout{** loaded for the language `#1'. Using the pattern for}%
\typeout{** the default language instead.}%
\else
\language=\csname l@#1\endcsname
\fi
#2}}
\providecommand{\BIBdecl}{\relax}
\BIBdecl

\bibitem{barbakh2008online}
W.~Barbakh and C.~Fyfe, ``Online clustering algorithms,'' \emph{International Journal of Neural Systems}, vol.~18, no.~03, pp. 185--194, 2008.

\bibitem{Vincent2021OKmeans}
V.~Cohen-Addad, G.~Benjamin, K.~Varun, and R.~Guy, ``Online k-means clustering,'' in \emph{International Conference on Artificial Intelligence and Statistics}.\hskip 1em plus 0.5em minus 0.4em\relax PMLR, 2021, pp. 1126--1134.

\bibitem{hore2008online}
P.~Hore, L.~Hall, D.~Goldgof, and W.~Cheng, ``Online fuzzy c means,'' in \emph{NAFIPS 2008-2008 Annual Meeting of the North American Fuzzy Information Processing Society}.\hskip 1em plus 0.5em minus 0.4em\relax IEEE, 2008, pp. 1--5.

\bibitem{Hall2011OnlineFCM}
L.~O. Hall and D.~B. Goldgof, ``Convergence of the single-pass and online fuzzy c-means algorithms,'' \emph{IEEE Transactions on Fuzzy Systems}, vol.~19, no.~4, pp. 792--794, 2011.

\bibitem{xu2015pca}
Q.~Xu, C.~Ding, J.~Liu, and B.~Luo, ``Pca-guided search for k-means,'' \emph{Pattern Recognition Letters}, vol.~54, pp. 50--55, 2015.

\bibitem{ashfahani2020devdan}
A.~Ashfahani, M.~Pratama, E.~Lughofer, and Y.-S. Ong, ``{DEVDAN:} deep evolving denoising autoencoder,'' \emph{Neurocomputing}, vol. 390, pp. 297--314, 2020.

\bibitem{ramasamy2020ORBM}
S.~Ramasamy, A.~Ambikapathi, and K.~Rajaraman, ``{Online RBM:} growing restricted boltzmann machine on the fly for unsupervised representation,'' \emph{Applied Soft Computing}, vol.~92, pp. 1--10, 2020.

\bibitem{graphRBM}
D.~Chen, J.~Lv, and Z.~Yi, ``Graph regularized restricted boltzmann machine,'' \emph{IEEE transactions on neural networks and learning systems}, vol.~29, no.~6, pp. 2651--2659, 2017.

\bibitem{hinton2007boltzmann}
G.~E. Hinton, ``Boltzmann machine,'' \emph{Scholarpedia}, vol.~2, no.~5, p. 1668, 2007.

\bibitem{baldi2012autoencoders}
P.~Baldi, ``Autoencoders, unsupervised learning, and deep architectures,'' in \emph{Proceedings of ICML workshop on unsupervised and transfer learning}.\hskip 1em plus 0.5em minus 0.4em\relax JMLR Workshop and Conference Proceedings, 2012, pp. 37--49.

\bibitem{vincent2008extracting}
P.~Vincent, H.~Larochelle, Y.~Bengio, and P.-A. Manzagol, ``Extracting and composing robust features with denoising autoencoders,'' in \emph{Proceedings of the 25th international conference on Machine learning}, 2008, pp. 1096--1103.

\bibitem{fernandez2023disentangling}
J.~Fernandez-de Cossio-Diaz, S.~Cocco, and R.~Monasson, ``Disentangling representations in restricted boltzmann machines without adversaries,'' \emph{Physical Review X}, vol.~13, no.~2, p. 021003, 2023.

\bibitem{gu2021refinements}
L.~Gu, L.~Yang, and F.~Zhou, ``Refinements of approximation results of conditional restricted boltzmann machines,'' \emph{IEEE Transactions on Neural Networks and Learning Systems}, vol.~34, no.~3, pp. 1228--1242, 2021.

\bibitem{roder2020energy}
M.~Roder, G.~H. de~Rosa, V.~H.~C. de~Albuquerque, A.~L. Rossi, and J.~P. Papa, ``Energy-based dropout in restricted boltzmann machines: why not go random,'' \emph{IEEE Transactions on Emerging Topics in Computational Intelligence}, vol.~6, no.~2, pp. 276--286, 2020.

\bibitem{golab2003issue}
L.~Golab and M.~T. {\"O}zsu, ``Issues in data stream management,'' \emph{ACM Sigmod Record}, vol.~32, no.~2, pp. 5--14, 2003.

\bibitem{Datastreams1}
S.~Guha, N.~Mishra, R.~Motwani, and L.~O'Callaghan, ``Clustering data streams,'' in \emph{Proceedings 41st Annual Symposium on Foundations of Computer Science}, 2000, pp. 359--366.

\bibitem{babu2016meta}
G.~S. Babu, X.-L. Li, and S.~Suresh, ``Meta-cognitive regression neural network for function approximation: application to remaining useful life estimation,'' in \emph{2016 International Joint Conference on Neural Networks (IJCNN)}.\hskip 1em plus 0.5em minus 0.4em\relax IEEE, 2016, pp. 4803--4810.

\bibitem{Senthil2022metacog}
J.~Senthilnath, K.~Abhishek, J.~Anurag, H.~K., T.~Meenakumari, S.~S., A.~Gautham, and J.~A. Benediktsson, ``{BS-McL:} bilevel segmentation framework with metacognitive learning for detection of the power lines in uav imagery,'' \emph{IEEE Transactions on Geoscience and Remote Sensing}, vol.~60, pp. 1--12, 2022.

\bibitem{ashfahani2021biasvar}
A.~Andri, P.~Mahardhika, L.~Edwin, and K.~Y. Yapp, ``Autonomous deep quality monitoring in streaming environments,'' in \emph{Proceedings of the international joint conference on neural networks}, 2021, pp. 1--8.

\bibitem{onlineRBM2016chen}
G.~Chen, X.~Ran, and S.~Sargur~N., ``Sequential labeling with online deep learning: Exploring model initialization,'' in \emph{Joint European Conference on Machine Learning and Knowledge Discovery in Databases}, 2016, pp. 772--788.

\bibitem{onlineAE2012Zhou}
G.~Zhou, S.~Kihyuk, and L.~Honglak, ``Online incremental feature learning with denoising autoencoders,'' in \emph{Artificial intelligence and statistics}, 2012, pp. 1453--1461.

\bibitem{cormode2005s}
G.~Cormode and S.~Muthukrishnan, ``What's hot and what's not: tracking most frequent items dynamically,'' \emph{ACM Transactions on Database Systems (TODS)}, vol.~30, no.~1, pp. 249--278, 2005.

\bibitem{datar2002estimating}
M.~Datar and S.~Muthukrishnan, ``Estimating rarity and similarity over data stream windows,'' in \emph{European Symposium on Algorithms}.\hskip 1em plus 0.5em minus 0.4em\relax Springer, 2002, pp. 323--335.

\bibitem{yuan2018clustering}
L.~Yuan, X.~Xiao, F.~Li, and N.~Deng, ``Clustering model based on rbm encoding in big data,'' in \emph{International Conference on Cloud Computing and Security}.\hskip 1em plus 0.5em minus 0.4em\relax Springer, 2018, pp. 387--396.

\bibitem{Senthil2024rbm}
J.~Senthilnath, G.~Nagaraj, S.~C. Sumanth, K.~Sushant, T.~Meenakumari, M.~Indiramma, and J.~A. Benediktsson, ``{DRBM-ClustNet}: A deep restricted boltzmann-kohonen architecture for data clustering,'' \emph{IEEE Transactions on Neural Networks and Learning Systems}, vol.~35, no.~2, pp. 2560--2574, 2024.

\bibitem{murphy2012machine}
K.~P. Murphy, \emph{Machine learning: a probabilistic perspective}.\hskip 1em plus 0.5em minus 0.4em\relax MIT press, 2012.

\bibitem{hinton2002training}
G.~E. Hinton, ``Training products of experts by minimizing contrastive divergence,'' \emph{Neural computation}, vol.~14, no.~8, pp. 1771--1800, 2002.

\bibitem{hinton2012practical}
------, ``A practical guide to training restricted boltzmann machines,'' in \emph{Neural networks: Tricks of the trade}.\hskip 1em plus 0.5em minus 0.4em\relax Springer, 2012, pp. 599--619.

\bibitem{Andri2022Clust}
A.~Ashfahani and P.~Mahardhika, ``Unsupervised continual learning in streaming environments,'' \emph{IEEE Transactions on Neural Networks and Learning Systems}, vol.~34, no.~12, pp. 9992--10\,003, 2023.

\bibitem{dogruparmak2014using}
S.~C. Dogruparmak, G.~A. Keskin, S.~Yaman, and A.~Alkan, ``Using principal component analysis and fuzzy c--means clustering for the assessment of air quality monitoring,'' \emph{Atmospheric Pollution Research}, vol.~5, no.~4, pp. 656--663, 2014.

\bibitem{lopez2004principal}
E.~L{\'o}pez-Rubio, J.~Munoz-P{\'e}rez, and J.~A. G{\'o}mez-Ruiz, ``A principal components analysis self-organizing map,'' \emph{Neural Networks}, vol.~17, no.~2, pp. 261--270, 2004.

\bibitem{yang2015deep}
Q.~Yang, H.~Wang, T.~Li, and Y.~Yang, ``Deep belief networks oriented clustering,'' in \emph{2015 10th International Conference on Intelligent Systems and Knowledge Engineering (ISKE)}.\hskip 1em plus 0.5em minus 0.4em\relax IEEE, 2015, pp. 58--65.

\bibitem{LCun1998MNIST}
L.~Yann, L.~Bottou, Y.~Bengio, and P.~Haffner, ``Gradient-based learning applied to document recognition,'' \emph{Proceedings of the IEEE}, vol.~86, no.~11, pp. 2278--2324, 1998.

\bibitem{xiao2017fashion}
H.~Xiao, K.~Rasul, and R.~Vollgraf, ``Fashion-mnist: a novel image dataset for benchmarking machine learning algorithms,'' \emph{arXiv preprint arXiv:1708.07747}, 2017.

\bibitem{Tarin2018KMNIST}
T.~Clanuwat, B.-I. Mikel, K.~Asanobu, L.~Alex, Y.~Kazuaki, and H.~David, ``Deep learning for classical japanese literature,'' \emph{arXiv preprint 1812.01718}, pp. 1--8, 2018.

\bibitem{nene1996columbia}
S.~A. Nene, S.~K. Nayar, H.~Murase \emph{et~al.}, ``Columbia object image library (coil-20),'' 1996.

\bibitem{wang2020deformable}
J.~Wang, C.~Xu, Z.~Yang, J.~Zhang, and X.~Li, ``Deformable convolutional networks for efficient mixed-type wafer defect pattern recognition,'' \emph{IEEE Transactions on Semiconductor Manufacturing}, vol.~33, no.~4, pp. 587--596, 2020.

\bibitem{lan2017deep}
W.~Lan, Q.~Li, N.~Yu, Q.~Wang, S.~Jia, and K.~Li, ``The deep belief and self-organizing neural network as a semi-supervised classification method for hyperspectral data,'' \emph{Applied Sciences}, vol.~7, no.~12, p. 1212, 2017.

\bibitem{PutityNMIequation}
X.~Liu, H.-M. Cheng, and Z.-Y. Zhang, ``Evaluation of community detection methods,'' \emph{IEEE Transactions on Knowledge and Data Engineering}, vol.~32, no.~9, pp. 1736--1746, 2020.

\end{thebibliography}

\vfill

\end{document}